\pdfoutput=1
\documentclass[11pt]{article}

\usepackage[]{acl}

\usepackage{times}
\usepackage{latexsym}
\usepackage[T1]{fontenc}
\usepackage[utf8]{inputenc}

\usepackage{microtype}
\usepackage{amsfonts} 
\usepackage{pifont}
\usepackage{subfigure}
\usepackage{cancel}
\usepackage{soul}
\usepackage{xcolor}
\usepackage{algorithm}
\usepackage{algorithmic}
\usepackage{float}
\usepackage{booktabs}
\usepackage{xspace}
\usepackage{mathrsfs}
\usepackage{multirow}
\usepackage{amsmath}
\usepackage{colortbl}
\usepackage{graphicx}
\newcommand{\ie}{\textit{i.e.,}\xspace}
\definecolor{bleudefrance}{rgb}{0.19, 0.55, 0.91}
\definecolor{yes}{RGB}{239,211,69}
\definecolor{carminered}{rgb}{1.0, 0.0, 0.22}
\definecolor{crimsonglory}{rgb}{0.75, 0.0, 0.2}
\newcommand{\hlc}[2][yellow]{{%
    \colorlet{foo}{#1}%
    \sethlcolor{foo}\hl{#2}}%
}

\title{Improving the Robustness of Summarization Systems with\\ Dual Augmentation}

\author{Xiuying Chen$^{1}$, Guodong Long$^{2}$, Chongyang Tao$^{3\dagger}$, Mingzhe Li$^{4}$, \\ \textbf{Xin Gao}$^{1\dagger}$, \textbf{Chengqi Zhang}$^{2}$,  \textbf{Xiangliang Zhang}$^{5,1\dagger}$\thanks{$\dagger$ Corresponding author.}\\
\\
$^1$Computational Bioscience Reseach Center, KAUST \\
$^2$AAII, School of CS, FEIT, University of Technology Sydney\\
$^3$Microsoft 
$^4$Ant Group 
$^5$University of Notre Dame\\
\texttt{xiuying.chen@kaust.edu.sa}
}

\renewcommand\footnotemark{}
\begin{document}
\maketitle
\begin{abstract}

A robust summarization system should be able to capture the gist of the document, regardless of the specific word choices or noise in the input.
In this work, we first explore the summarization models' robustness against perturbations including word-level synonym substitution and noise.
To create semantic-consistent substitutes, we propose a SummAttacker, which is an efficient approach to generating adversarial samples based on language models.
Experimental results show that state-of-the-art summarization models have a significant decrease in performance on adversarial and noisy test sets.
\textcolor{black}{Next, we analyze the vulnerability of the summarization systems and explore improving the robustness by data augmentation.
Specifically, the first brittleness factor we found is the poor understanding of infrequent words in the input.
Correspondingly, we feed the encoder with more diverse cases created by SummAttacker in the input space.
The other factor is in the latent space, where the attacked inputs bring more variations to the hidden states.
Hence, we construct adversarial decoder input and devise manifold softmixing operation in hidden space to introduce more diversity.
}
Experimental results on Gigaword and CNN/DM datasets demonstrate that our approach achieves significant improvements over strong baselines and exhibits higher robustness on noisy, attacked, and clean datasets\footnote{\url{https://github.com/iriscxy/robustness}}.
\end{abstract}

\section{Introduction}

\begin{table}[htb]
\small
\begin{tabular}{c|l}
\toprule
\multicolumn{2}{c}{Perturbation class: \textit{Typo}}\\ \hline
 Input & ministers from the european union \\&and its mediterranean neighbors gathered \\&here under heavy security on monday \\&for an unprecedented \\&\hlc[bleudefrance!40]{conference ($\rightarrow$confecence)} \\&on economic and political cooperation .\\
        \hline
 
 Original  & eu mediterranean nations meet for first-ever \\ Summary& conference on cooperation. $\checkmark$\\
 \hline
 Perturbed    &eu mediterranean ministers meet in \\
 Summary      & greece under heavy security.  $\mathbf{\times}$\\
\midrule

\multicolumn{2}{c}{Perturbation class: \textit{Synonym substitution}}\\ \hline
 Input & judge leonie brinkema ordered september \\& \#\# conspirator zacarias moussaoui removed \\& from the court here on monday \\& after he repeatedly rejected his court- \\&  appointed defense \hlc[bleudefrance!40]{attorney ($\rightarrow$barrister)} . \\
        \hline
 
 Original  &moussaoui removed from court    \\Summary & after rejecting defense attorneys. $\checkmark$\\
 \hline
 Perturbed    &moussaoui removed from court   \\
 Summary      & after rejecting defense barris.  $\mathbf{\times}$
  \\
\midrule

 Input &  president barack obama is imploring \\& voters to support his  \hlc[bleudefrance!40]{government ($\rightarrow$party)} 's \\& economic policies even though he \\& acknowledged that those policies haven't \\& brought about a recovery less than two \\& months before the midterm elections .\\
        \hline
 
 Original  & obama says voters should back his \\ Summary& economic policies.$\checkmark$\\
 \hline
 Perturbed    &obama urges voters to back gop  \\
 Summary      &  economic policies. $\mathbf{\times}$\\
\bottomrule
\end{tabular}
\caption{\textcolor{black}{Examples of vulnerability to BART-based summarization model. 
All examples show an initially correct summary turning into a wrong summary due to  small changes in the input, \textit{e.g.,} mis-spelling and synonym substitution.}}
\vspace{-5mm}
\label{table:wrong_translation} 
\end{table}

Humans have robust summarization processing systems that can easily understand diverse expressions and various wording, and overcome typos, misspellings, and the complete omission of letters when reading \cite{rawlinson2007significance}.
However, studies reveal that small changes in the input can lead to significant performance drops and fool state-of-the-art neural networks \cite{goodfellow2014explaining,belinkov2017synthetic,cheng2018towards}.
In text generation fields such as machine translation,
\citet{belinkov2017synthetic} showed that state-of-the-art models fail to translate even moderately noisy texts,
\citet{cheng2018towards} found that the generated translation is completely distorted by only replacing a source word with its synonym.
However, the robustness on summarization models is less explored.
Here, we show three summarization examples from the Gigaword dataset in Table~\ref{table:wrong_translation}.
A fine-tuned BART model will generate a worse summary for a minor change in the input including misspelling errors and synonym substitution, which often happen in practice due to the carelessness and habit of word usage in writing. 
Take the second case for example, an English user and an American user who use  \emph{barrister} or \emph{attorney} will obtain summaries of different qualities. 
In the third case, a synonym word replacement even changes the subject of canvassing.
Such weakness of summarization systems can lead to serious consequences in practice.

Despite its importance, robustness in summarization has been less explored.
\citet{jung2019earlier} and \citet{kryscinski2019neural} examined positional bias and layout bias in summarization.
\citet{liu2021noisy} introduced multiple noise signals in self-knowledge distillation to improve the performance of student models on benchmark datasets, but they did not explicitly evaluate the robustness of summarization models against noise.

Hence, in this work, we first evaluate the robustness of the existing state-of-the-art summarization systems against word-level perturbations including noise and adversarial attacks. 
The noise consists of natural human errors such as typos and misspellings.
To create the adversarial attack test set, we come up with a model named SummAttacker.
The core algorithm of SummAttacker is to find   vulnerable words in a given document for the target model and then apply language models to find substituted words adjacent in the opposite direction of the gradient to maximize perturbations.
We validate the effectiveness of SummAttacker on benchmark datasets with different attributes, \ie Gigaword and CNN/DailyMail. 
Experiment results show that by only attacking one word (1\% token) in Gigaword and 5\% tokens in CNN/DailyMail, the existing summarization models have drastically lower performance.

\textcolor{black}{We next conduct a vulnerability analysis and propose two corresponding solutions to improve robustness.}
% seek to improve the robustness of summarization models by data augmentation, aiming to introduce more diversity in the training process.
% One solution is data augmentation which introduces diversity in the training process.
\textcolor{black}{Our first conjecture is that worse summaries can be caused by replacing common words with uncommon and infrequently-used words, which the model might not understand well.
Hence, we employ the outputs from SummAttacker as inputs for the encoder, so as to improve the diversity in the discrete input  space.
The second influencing factor is that the attacked inputs introduce more variations in the latent space.
Correspondingly, we aim to expose the model to more diverse hidden states in the training process.
Specifically, we build soft pseudo tokens by multiplying the decoder output probability with target token embeddings.
These soft pseudo tokens and original tokens are then manifold softmixed on a randomly selected decoder layer to enlarge the training distribution.
% The manifold softmix operation interpolates the representations of the two inputs on a randomly selected decoder layer.
}
\textcolor{black}{
The  interpolations leveraged in deeper hidden layers help capture  higher-level information, improve semantic diversity, and provide additional training signal \cite{zeiler2014visualizing}.
% The virtual cases obtained by the softmix operation are centered around each training example, named as vicinity distribution, in the high-level semantic feature space. 
% The softmix operation thus leads to smoother decision boundaries, which are well-established factors of generalization \cite{bartlett1999generalization,verma2019manifold}.
}
Experiments show that our dual augmentation \textcolor{black}{for both encoder and decoder} improves the robustness of summarization models on noisy and attacked test datasets.
 
Our main contributions are as follows:

$\bullet$ We empirically evaluate the robustness of recent summarization models against perturbations including noise and synonym substitutions.
% Experiments show that these models have a significant decrease in performance towards both variations.

$\bullet$ To improve the robustness of summarization models, we propose a dual data augmentation method that introduces diversity in the input and latent semantic spaces.

$\bullet$ Experimental results demonstrate that our augmentation method brings substantial improvements over state-of-the-art baselines on benchmark datasets and attacked test datasets.

\section{Related Work}
We discuss related work on robust abstractive summarization, adversarial examples generation, and data augmentation.

\paragraph{Robust Abstractive Summarization.}
Ideally, a robust text generation system should consistently have high performance even with small perturbations in the input, such as token and character swapping \cite{jin2019bert}, paraphrasing \cite{gan2019improving}, and semantically equivalent adversarial rules \cite{ribeiro2018semantically}.
% In the text generation field, several approaches have been tried to improve the robustness of machine translation.
Considerable efforts have been made in the text generation field.
For example, 
% \citet{belinkov2017synthetic} proposed structure-invariant word representations and training on noisy texts to improve robustness.
\citet{cheng2019robust} defended a translation model with adversarial source examples and target inputs.
% For the question answering task, \citet{gan2019improving} released two paraphrased test sets for evaluation of QA models' robustness to paraphrasing.
However, the robustness in the summarization task has been less explored.
\citet{jung2019earlier} and \citet{kryscinski2019neural} showed that summarization models often overfit to positional and layout bias, respectively.
% , which reflect the robustness of summarization models under distribution shift.
% \citet{aghajanyan2020better} proposed to learn robust representations through regularized fine-tuning, and verify its effectiveness on summarization tasks.
In contrast, in this work, we focus on the robustness of summarization models against word-level perturbations.

% In recent years, text generation has made impressive progress~\cite{Pan2019ImprovingOD,stiennon2020learning,ma2021global}, which promotes the progress of abstractive summarization.
% Abtractive summarization task generates novel words and phrases not featured in the source text to capture the salient ideas of the source text~\cite{liu2018generative}.
% Most works apply encoder-decoder architecture to implicitly learn the summarization procedure \cite{gehrmann2018bottom,celikyilmaz2018deep}.
% More recently, applying pretrained language models as encoder \cite{Liu2019TextSW,zhou2021entity} or pre-training the generation process leveraging a large-scale of unlabeled corpus \cite{zhangpegasus,lewis2020bart} are shown to bring significant improvements .
% Explicit structures modeling has also been shown to be effective in summarization task.
% For example, \citet{jin2020semsum} incorporated semantic dependency graphs to help generate sentences with better semantic relevance, and \citet{wu2021bass} came up with a unified semantic graph to aggregate relevant disjoint context from the input.

\paragraph{Adversarial Examples Generation.}
Classic attacks for text usually adopt heuristic rules to modify the characters of a word \cite{belinkov2017synthetic} or substitute words with synonyms \cite{ren2019generating}.
These heuristic replacement strategies make it challenging to find optimal solutions in the massive space of possible replacements while preserving semantic consistency and language fluency.
Recently, 
% pre-trained language models have been used for attacks with more diverse and fluent candidates.
\citet{li2020bert} proposed to generate adversarial samples for the text classification task using pre-trained masked language models exemplified by BERT.
In this paper, we focus on attacking summarization models, which is a more challenging task, since the model compresses the input, and perturbations on unimportant parts of the source might be ignored.

\paragraph{Data Augmentation.} 
Data augmentation aims to generate more training examples without incurring additional efforts of manual labeling, which can improve the robustness or performance of a target model.
Conventional approaches introduce discrete noise by adding, deleting, and/or replacing characters or words in the input sentences \cite{belinkov2017synthetic}.
More recently, continuous augmentation methods  have been proposed.
\citet{cheng2020advaug} generated adversarial sentences from a smooth interpolated embedding space centered around observed training sentence pairs, and shows its effectiveness on benchmark and noisy translation datasets.
\citet{xie2021target} proposed a target-side augmentation method, which uses the decoder output probability distributions as soft indicators.
\citet{chen2023learning} selectively augmented training dataset considering representativeness and generation quality.
In this work, we propose a dual augmentation method that utilizes discrete and virtual augmented cases.

% \section{Problem Formulation}
% \label{sec:formulation}

% Before presenting our approach for the faithful summarization task, we first introduce our notations and key concepts. 

\section{The Proposed SummAttacker}
\begin{figure}[t]
\centering
\includegraphics[scale=0.52]{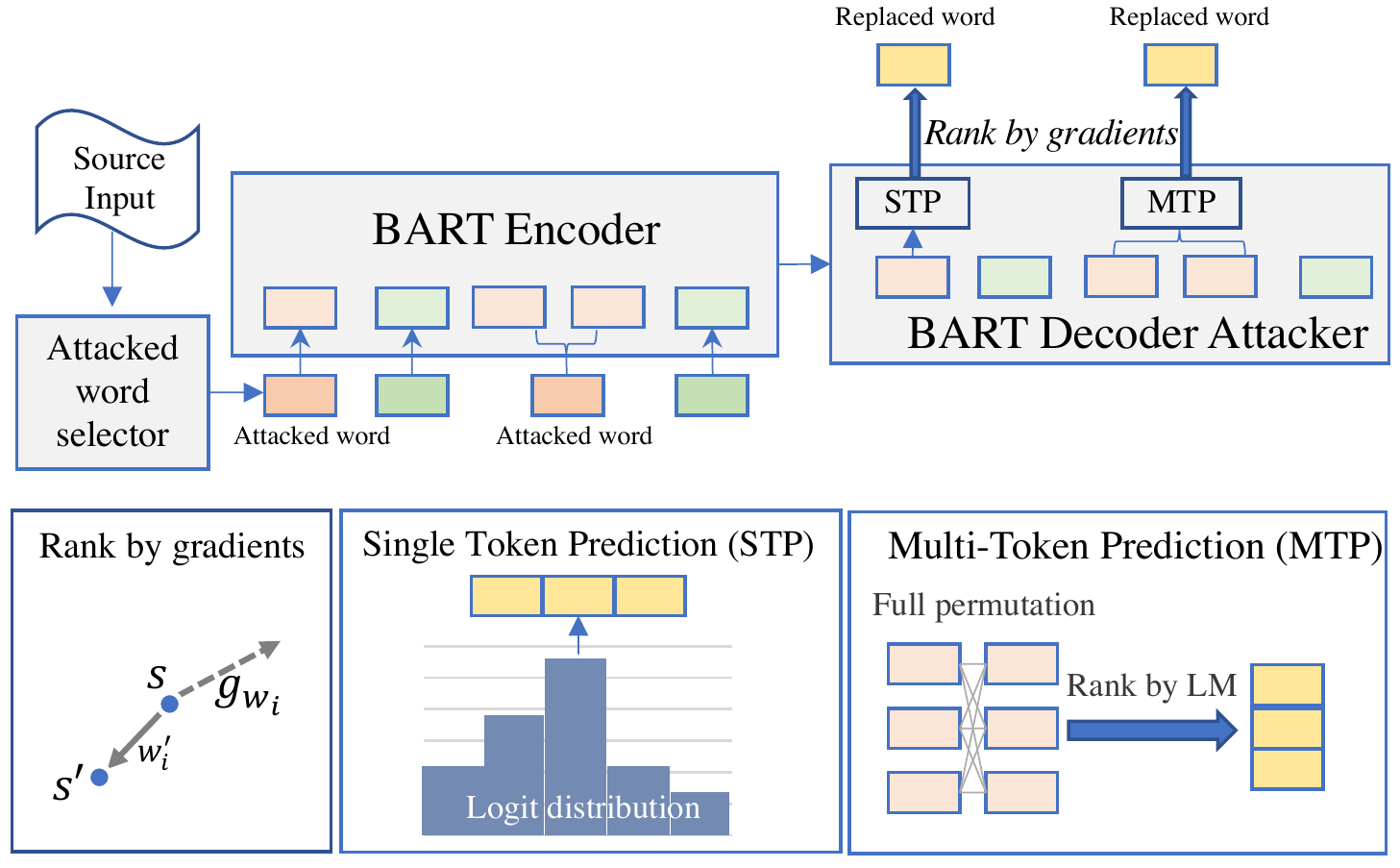}
\caption{
    Overview of SummAttacker. It first selects vulnerable words to attack, and then replaces them with words based on language model (LM) prediction and gradient-based ranking.
   The replacement word $w_{i}^{\prime}$  changes the model state $s$ to $s'$ in the opposite direction of optimization,  $-\mathbf{g}_{w_i}$.}
\label{fig:attack}
\end{figure}

\textcolor{black}{
%% the problem is re-defined
Formally, given a trained summarization model with parameters $\boldsymbol{\theta}$, the purpose of an attacking model is to slightly perturb the input $x$ such that the summarization output of the perturbed $\hat{x}$ deviates away from the target summary $y$:
} 
\begin{equation}
\resizebox{.82\hsize}{!}{$\{\hat{x} | \mathcal{R}\left(\hat{x}, x\right) \leq \epsilon, \underset{\hat{x}}{\operatorname{argmax}}-\log P\left(y| \hat{x} ; \boldsymbol{\theta}\right)\}$},
% \vspace{-0.25cm}
 \label{attack_form} 
\end{equation} 
where $\mathcal{R}\left(\hat{x}, x\right)$ captures the degree of imperceptibility for a perturbation, \textit{e.g.,} the number of perturbed words. To make a maximal impact on the summarization output  with a perturbation budget $\epsilon$, a classical way is to launch gradient-based attacks \cite{cheng2019robust}.
%By choosing to replace a word in the opposite direction of the optimization gradient, we aim to perturb the summarization model as much as possible.
In this section, we propose a SummAttacker 
\textcolor{black}{for crafting adversarial samples that may differ only a few words  from genuine inputs but have low-quality summarization results. Due to its capacity and popularity, we take  BART \cite{lewis2020bart}  as the backbone summarization model, as shown in Fig.\ref{fig:attack}.  } 
%based on BART \cite{lewis2020bart} and gradients to craft adversarial samples to fool the summarization model, as shown in Figure~\ref{fig:attack}.
% Our method consists of two steps: (1) selecting the vulnerable words for attack and then (2) replacing them with semantically similar and grammatically correct words based on language model gradient comparison.

\textbf{Attacked Word Selector.}
Since it is intractable to obtain an exact solution for Equation~\ref{attack_form}, we, therefore, resort to a greedy approach to circumvent it. 
% Firstly, we find the most important word to the summarization model based on the attention mechanism.
% Then, we use pretrained language models to predict the replacing word.
\textcolor{black}{In BART kind of  summarization model based on Transformer architecture,}
%Here we employ the popular Transformer architecture as the summarization model.
%The 
% an encoder first transforms an input $x$ to a sequence of vector representations $\mathbf{s}$, from which the decoder generates an output sequence $\hat{y}$.
% Multi-head attention enables prediction at each step to attend to overall inputs from multiple representation subspaces in the encoder.
 \textcolor{black}{ the sequence representation vector $\boldsymbol{s}$ of input tokens in $x$ } is first projected   to keys $\boldsymbol{K}$ and values $\boldsymbol{V}$ using different linear mapping functions. % projections. 
At the $t$-th decoding step, the   hidden state of the previous decoder layer is projected to the query vector $\boldsymbol{q}_t$. 
Then   $\boldsymbol{q}_t$ is  multiplied by keys $\boldsymbol{K}$ to obtain an \textcolor{black}{attention score}  $\boldsymbol{a}_{t}$ and the $t$-th decoding output: 
% which is used to calculate a weighted sum of values $\mathbf{V}$ \textcolor{black}{ for predicting the $t$-th decoding output:} 
\begin{equation*} 
\resizebox{.85\hsize}{!}{$\operatorname{Attn}\left(\boldsymbol{q}_{t}, \boldsymbol{K}, \boldsymbol{V}\right)=\boldsymbol{a}_{t} * \boldsymbol{V} , \;
\boldsymbol{a}_{t}=\operatorname{softmax}\left(\frac{\boldsymbol{q}_{t} \boldsymbol{K}^{T}}{\sqrt{d}}\right),$}
 \label{attack} 
\end{equation*}
%     	\begin{align}
% 	\alpha_{i, j} =\frac{\exp \left(Q_{i,j} K_{i,j}\right)}{\sum_{n=1}^{L_r} \exp \left(Q_{i,n} K_{i,n}\right)}, 
% 	a_{i} =\sum_{j=1}^{L_r} \frac{\alpha_{i, j} V_{i,j}}{\sqrt {d_e}}, 
% 	\end{align}
where $d$ is the hidden dimension.
\textcolor{black}{A token that obtains the highest attention score over all decoding steps is the most important and influential one to the summarization model. }
%The token that obtains the highest attention score over all steps is chosen as the attacked word, as the summarization model relies on it the most.
% Note that BART uses Bytes-Pair-Encoding (BPE), hence, one token may correspond to one word or a subword. 
\textcolor{black}{ We select the word $w_i$ to attack if it contains or equals the most important token. To avoid   changing factual information, we restrict $w_i$ not to be people names and locations. 
}
%Here we denote the chosen word that contains or equals to the most important token as $w_i$.
%Note that we do not select people name and location to avoid changing factual information in the input.

\textbf{Attacking with LM and Gradients.}
% After deciding on the word $w_i$ to attack, we need to find a replacement word that is semantically similar to $w_i$ but is adversarial to the summarization model.
Next, we aim to find a replacement word that is semantically similar to $w_i$ but is adversarial to the summarization model.
Language models are empowered to generate sentences that are semantically accurate, fluent, and grammatically correct. 
We take advantage of this characteristic to find a replacement word $w'_i$ for the target word $w_i$. 
The general idea is  to first identify the top likely candidates that are predicted by the language model for $w_i$, and then select the best candidate with the guidance of  prediction gradient. 
%We first apply BART to predict the possible words that are similar to the attacked word.
%The characteristics of language model ensure that the generated sentences are semantically accurate, fluent, and grammatically correct.
% We do not mask the selected word $w_i$ but instead, use the original sequence as input to produce more semantically accurate replacements \cite{zhou2019bert}.
% This can also speed up the generation process as we do not need to iteratively run the BART model for each masked word.

Concretely, we first feed the tokenized sequence into the BART model to get a prediction for the attacked word $w_i$. 
% We take the most possible $K$ predictions for the attacked word as candidates, denoted as $\mathcal{V}_K$.
As shown in Fig.\ref{fig:attack}, for $w_i$ with a single token, we use STP (Single Token Prediction) operation to simply obtain the top $K$ predictions that are semantically similar to $w_i$.
For $w_i$ with multiple tokens, we have MTP (Multi-Token Prediction), which lists $c \times K$ possible combinations from the prediction, where $c$ is the token number in the word.
Then we rank the perplexity of all combinations to get the top-$K$ candidate combinations, denoted as $\mathcal{V}_K$.
We filter out stop words and antonyms using NLTK and synonym dictionaries.

\textcolor{black}{Following the idea of a gradient-based attack model, we then find the most adversarial word $w'_i$ that deviates from $w_i$ towards a change aligned with the prediction gradient:}
%Finally, we use a greedy approach based on the gradient to find the most adversarial word $w'_i$ from the $K$ candidates for the word $w_i$:
\begin{align} \footnotesize
    \begin{aligned} 
    \mathbf{g}_{w_{i}} &=\nabla_{\mathbf{e}\left(w_{i}\right)}\log P(y | x ; \boldsymbol{\theta}),\\
w_{i}^{\prime} &=\underset{w \in \mathcal{V}_{K}}{\operatorname{argmax}} \; \operatorname{sim}\left(\mathbf{e}(w)-\mathbf{e}\left(w_{i}\right), -\mathbf{g}_{w_{i}}\right),
\end{aligned}
\end{align}
where sim(·, ·) is cosine distance, and $\mathbf{e}$ is   word embedding function.
\textcolor{black}{As shown in Fig.~\ref{fig:attack}, the replacement word $w_{i}^{\prime}$  changes the model state $s$ to $s'$ in the opposite direction of optimization,  $-\mathbf{g}_{w_i}$.}

\section{Dual Augmentation}

\begin{figure}[tb]
\centering
\includegraphics[scale=0.485]{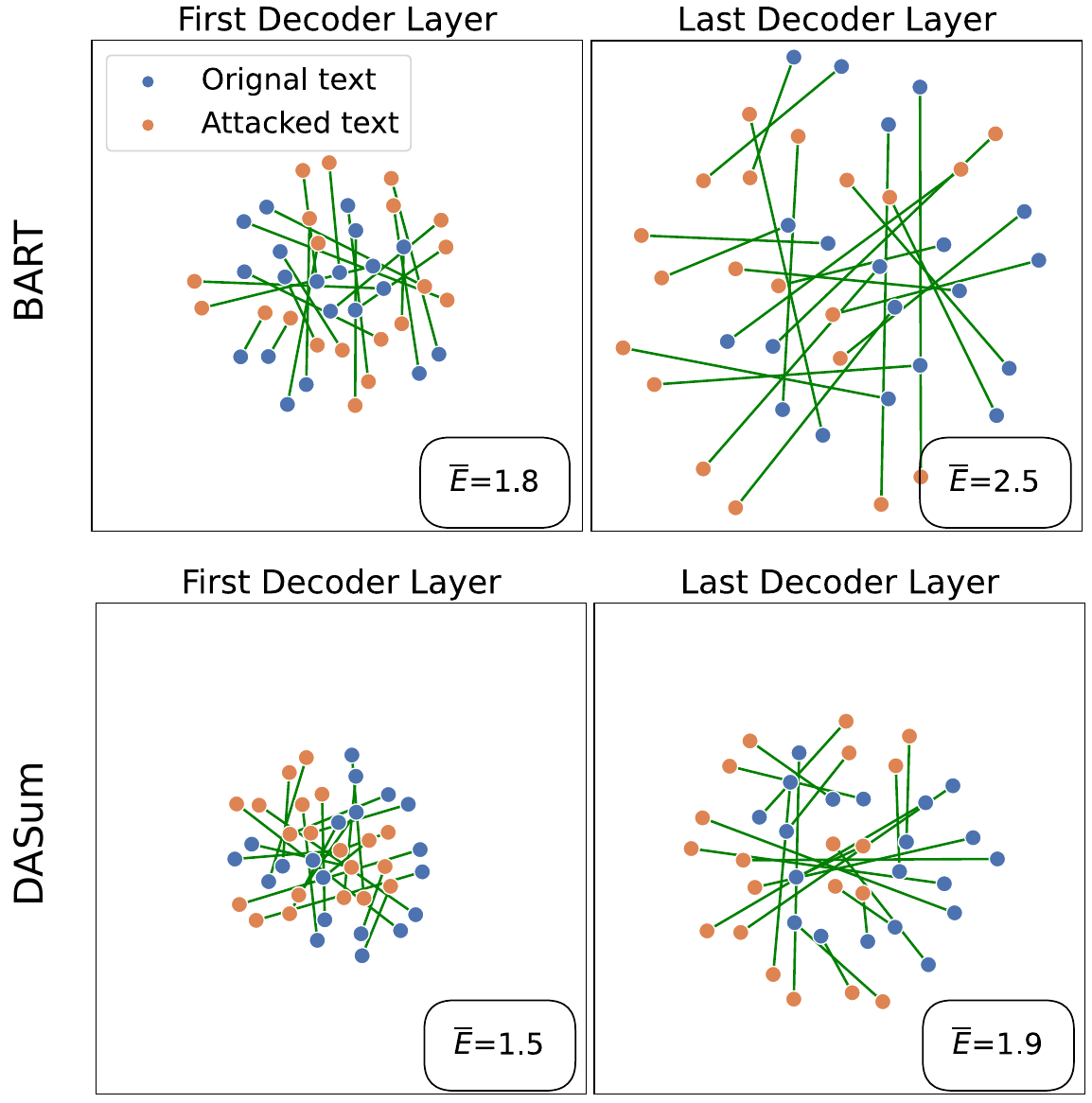}
\caption{
    t-SNE visualization of the hidden states in BART and our DASum, when taking original and attacked inputs. 
    $\overline{E}$ is the average Euclidean distance of paired original and attacked  states before using t-SNE.
}
\label{f1}
\end{figure}

\begin{figure*}
\centering
\includegraphics[scale=0.5]{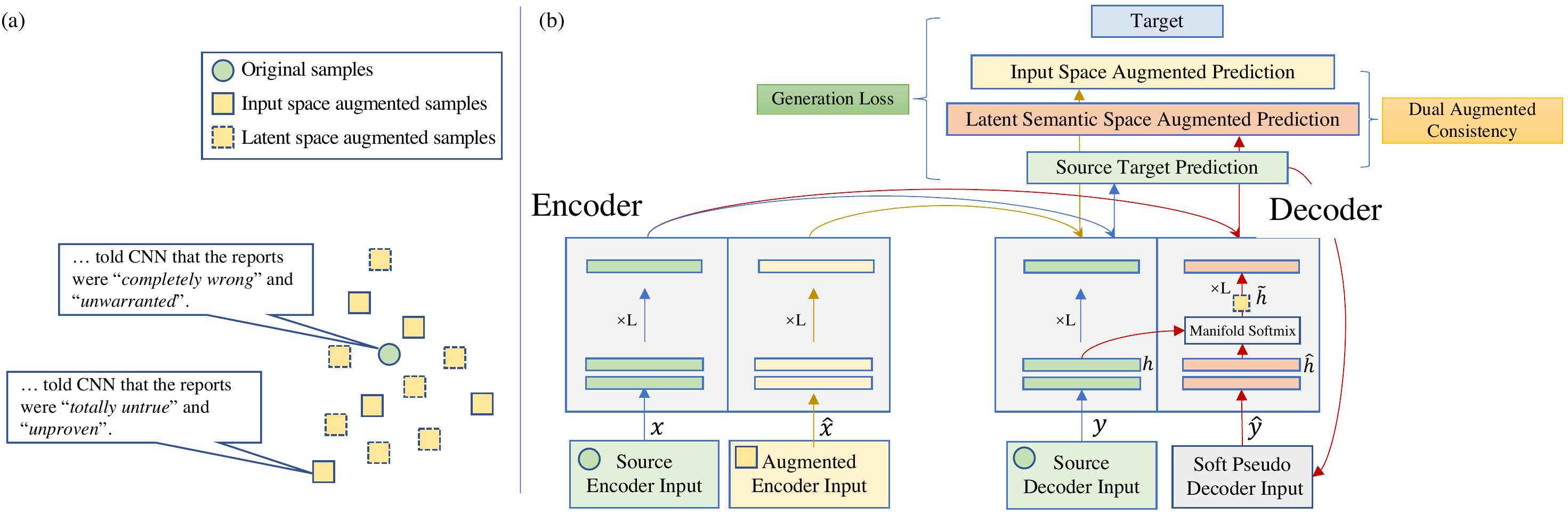}
\caption{
    (a) Illustration of training examples sampled from vicinity distributions that could cover variants of literal expression under the same meaning.
    (b) The architecture of our dual data augmentation approach.
    % The $\bigotimes$ denotes mixup operation.
}
\label{fig:aug}
\end{figure*}

\label{dual}

\textcolor{black}{
With the proposed  attacking model, we first analyze the  influences of  attacking, and then propose our DASum to counter the negative effects.
% By experiments in \S~\ref{robustevaluate}, we find that the performance of existing summarization models drop greatly on perturbed datasets.
% Hence, we are interested in the factors that lead to such vulnerability, so as to improve robustness.
}

\textcolor{black}{ \textbf{Vulnerability Analysis.} 
We first look into the word perturbation in attacked inputs that result in worse summaries. Our conjecture is that  worse summaries can be caused by replacing common words with uncommon and infrequently-used words, which the model might not understand  well. 
Through the analysis of 50 worse summary cases,  our conjecture is verified by the observation  that the frequency of the replacement words is 4 times lower than the original words on average. 
Especially for those worse summaries including unexpected words not existing in the  input, we found that the co-occurrence of the unexpected word in the generated summary and the replacement word in the  input is usually high.
Take the third case with unexpected work \emph{gop} in Table~\ref{table:wrong_translation} for example, the co-occurrence for the word pair \{\emph{party}, \emph{gop}\} is 6 times higher than that of \{\emph{government}, \emph{gop}\}.
These analysis results imply that the model's vulnerability is highly related to the word frequency distribution and the diversity of the training documents.
}
% The above results show that the model's understanding of words or word pairs is highly related to their frequencies.}

% vulnerability of 

Next, we investigate the influence of attack in the latent space.
\textcolor{black}{It is well known that in the text generation process, a change of a predicted preceding word will influence the prediction of words after it, since the following prediction will attend to the previously generated words~\cite{lamb2016professor}.
This error accumulation problem can be more severe in attacked scenarios since the perturbations can bring more variety in the decoder space.}
To verify our assumption, we evaluate the change in  hidden states  of the BART model for 20 cases in the original and the corresponding attacked test sets. 
The top part of Fig.\ref{f1} visualizes the hidden states in the first and last BART decoder layer.
\textcolor{black}{It can be seen that as the information flows from the low  to high layers in the decoder, the hidden states in the latent space show larger diversity, as the distances between paired hidden states get larger.
We also calculate the Euclidean distance $\overline{E}$ of paired states, which increases from 1.8 to 2.5.
\textcolor{black}{To improve the summarization robustness against attacks, the 
decoder could be trained with augmentation in latent space to comfort with diversity.
%deviation of paired hidden states  should be reduced. 
}
%Hence, we assume that it is necessary to improve the robustness of the decoder towards diverse hidden states in the training stage.
}

\textcolor{black}{
\textbf{Augmentation Design. }
Based on the above analysis, we first propose to incorporate the corpus obtained by SummAttacker as  \emph{augmentation input for encoder}, so as to improve the diversity of words in training documents (illustrated as yellow squares with solid lines in Fig.\ref{fig:aug}(a)).
To alleviate the impact of perturbation on the decoding process,  we  propose a continuous data augmentation method in the \emph{latent space of decoder}, where multiple virtual representations are constructed for each training instance to make the decoder be exposed to diverse variants of the latent representation of the same input document (illustrated as yellow squares with dash lines in Fig.\ref{fig:aug}(a)). 
}

\textbf{Input Space Augmentation.}
The input space augmentation in the encoder side is straightforward, as the output from SummAttacker can be directly employed as encoder inputs.
Concretely, we use SummAttacker to automatically generate an augmented input document for the original document, denoted as $\hat{x}$.
We then train the summarization model with the original  and augmented dataset, where the training objective is denoted as $\mathcal{L}_o=\log P\left(y|x\right)$ and $\mathcal{L}_e=\log P\left(y| \hat{x}\right)$, respectively.
We also randomly add noisy words in both inputs.
We show this process in Fig.\ref{fig:aug}(b), where we draw the same encoder twice to denote the training on original and augmented  inputs.

\textbf{Latent Semantic Space Augmentation.}
Based on the vulnerability analysis in the decoding process, we are motivated to 
\textcolor{black}{mitigate the impact of adversarial attacks by exposing the decoder to diverse variants of the latent representations. 
The variants are established by an adversarial input and a manifold softmix technique applied on randomly selected layers in the decoder. 
}

We first define a virtual adversarial decoder input $\hat{y}_t$ apart from the original input $y_t$ by  integrating the embedding of words that are all likely to be generated. 
Let $\mathbf{l}_t$ be the decoder's predicted logits before softmax, where $t \in \{1, 2,..., m \}$, 
 $l_t[v]$ be the logit of $v$ token, and $m$ is the token length of $y$. We compute the pseudo decoder inputs as:
\begin{equation}
\footnotesize
    \hat{y}_{t}= \frac{\exp \left(\mathbf{l}_{t} / T\right)}{\sum_{v=1}^{|\mathcal{V}|} \exp \left(l_{t}[v] / T\right)} \mathbf{W},
    \label{temperature}
\end{equation}
where  $\mathcal{V}$ is the vocabulary size, $\mathbf{W}$ is the  word embedding matrix with size $|\mathcal{V}| \times d$,   $T$ is the softmax temperature.

Next, we  construct the virtual adversarial hidden states   in the decoder by interpolating $\boldsymbol{h}^{k}$ and $\boldsymbol{\hat{h}}^{k}$, which are the hidden states of inputs $\boldsymbol{y}$ and $\boldsymbol{\hat{y}}$ at a randomly selected $k$-th layer:
\begin{equation} \footnotesize
    \tilde{\boldsymbol{h}}^{k}=\lambda \boldsymbol{h}^{k}+(1-\lambda) \hat{\boldsymbol{h}}^{k},
    \label{ratio}
\end{equation}
where  $\lambda$ is the mixup ratio between 0 and 1.
The mixup layer $k \in [0, L]$, where $L$ is the decoder layer number.

\textcolor{black}{In the decoding process, 
$\hat{y}_{t}$ servers as variants of $y_t$ and  integrates the embedding of words that are likely to be generated in each step. 
The variants of hidden states $\tilde{\boldsymbol{h}}^{k}$ behave like the hidden states of attacked input text. 
The latent space augmentation objective is  $\mathcal{L}_d=\log P\left(y |x, \hat{y}\right)$.
As shown in   Fig.\ref{fig:aug}, the latent semantic space augmented prediction  is a kind of additional training task for decoder with variant samples indicated by yellow squares with dash lines. Note that our proposed manifold  softmix differs from the  target-side augmentation  in  \citet{xie2021target}, which mixed the pseudo decoder input with the ground truth input in the word embedding layer, and only introduces low-level token variations.
}

Lastly, according to recent studies \cite{chen2020simple},  maximizing the consistency across various augmented data that are produced from a single piece of data might enhance model performance.
Herein, we minimize the bidirectional Kullback-Leibler (KL) divergence between the augmented data and real data, to stabilize the training:
\begin{equation} \footnotesize
   \begin{aligned}
    \mathcal{L}_c&=\mathcal{D}_{KL}\left(P\left(y |x\right)\| P\left(y |x, \hat{y}\right)\right)\\&+\mathcal{D}_{KL}\left(P\left(y |x\right)\| P\left(y |\hat{x}\right)\right).
\end{aligned} 
\end{equation}

Our final loss function is defined as $\mathcal{L}_o+\mathcal{L}_e+\mathcal{L}_d+\mathcal{L}_c$.

\section{Experimental Setup}
\label{sec:experiment}

\subsection{Dataset}
We experiment on two public datasets, Gigaword \cite{napoles2012annotated} and CNN/DM \cite{hermann2015teaching}, which have been widely used in previous summarization works.
The input document in Gigaword contains 70 words, while CNN/DM consists of 700 words on average.
Hence, we can examine the effectiveness of our methods on datasets of different distributions.

\subsection{Comparison Methods}

Our baselines include the following models:

\noindent \textbf{BART} \cite{lewis2020bart} is a state-of-the-art abstractive summarization model pretrained with a denoising autoencoding objective.

\noindent \textbf{ProphetNet} \cite{qi2020prophetnet} is a pre-training model that introduces a self-supervised n-gram prediction task and n-stream self-attention mechanism.

\noindent \textbf{R3F} \cite{aghajanyan2020better} is a robust text generation method, which replaces adversarial objectives with parametric noise, thereby discouraging representation change during fine-tuning when possible without hurting performance.

\noindent \textbf{SSTIA} \cite{xie2021target} augments the dataset from the target side by mixing the augmented decoder inputs in the embedding layer.

\begin{figure}[tb]
\centering
\includegraphics[scale=0.25]{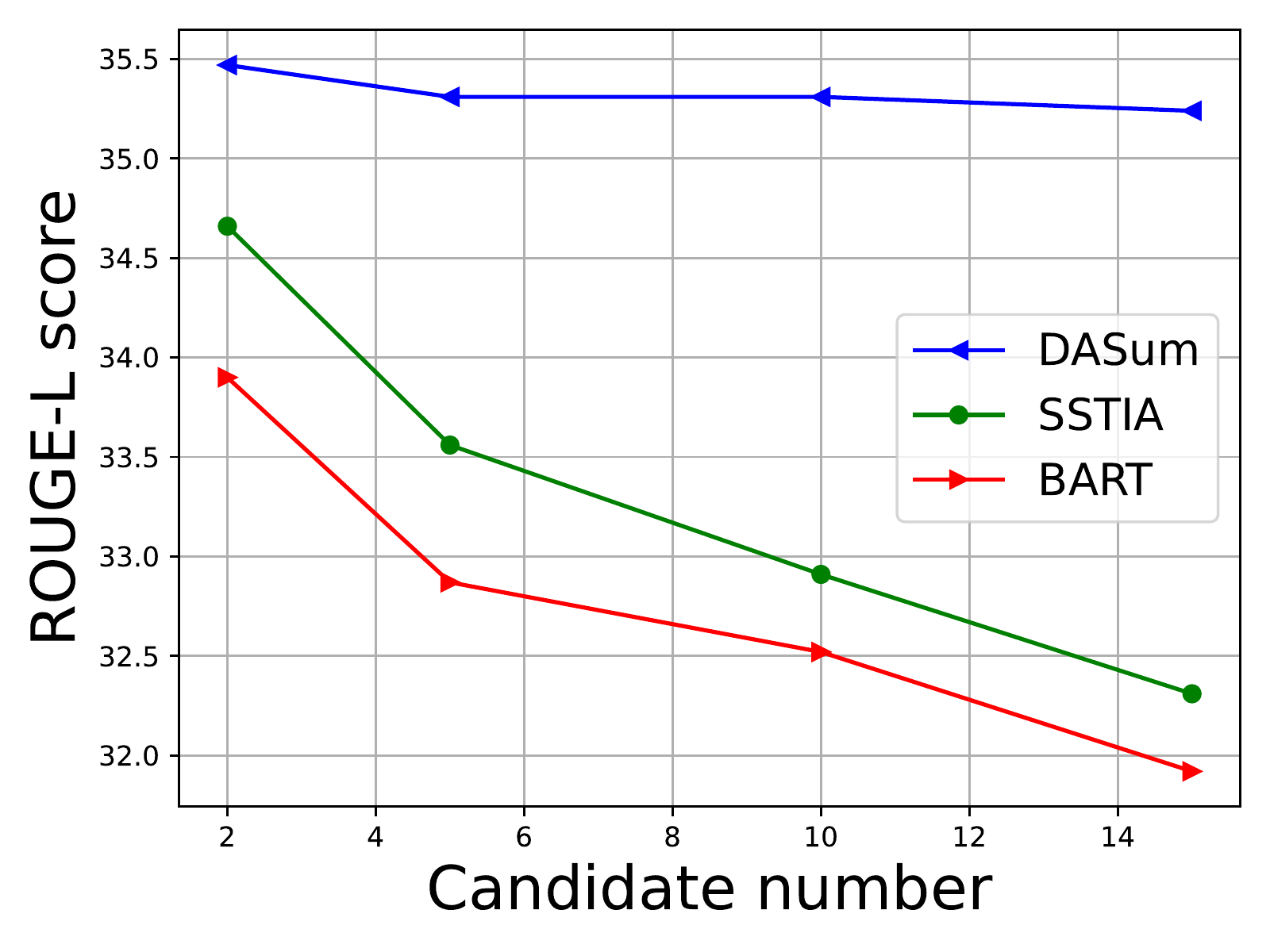}
\caption{
    Performance of different models on the Gigaword test set when attacked by SummAttacker with different candidate number $K$.
}

\label{fig:candidate}
\end{figure}

\subsection{Implementation Details}

We implement our experiments in Huggingface on NVIDIA A100 GPUs, and start finetuning based on pretrained models facebook/bart-large.
% We build our models based on BART and follow its hyperparameter settings.
Concretely, there are 12 encoding layers in the encoder and the decoder.
The activation functions are set to GeLUs and parameters are initialized from $\mathcal{N}(0,0.02)$.
We use Adam optimizer with $\epsilon$ as 1e-8 and $\beta$ as (0.9, 0.98). 
We used label smoothing of value 0.1, which is the same value as \citet{vaswani2017attention}.
Then attacking candidate number $K$ is set to 10 based on the parameter study.
The learning rate is set to 3e-5.
The warm-up is set to 500 steps for CNN/DM and 5000 for Gigaword. 
The batch size is set to 128 with gradient accumulation steps of 2.
Following \citet{xie2021target}, the temperature in Equation~\ref{temperature} is set to 0.1 for CNN/DM and 1 for Gigaword, and the mixup ratio $\lambda$ in Equation~\ref{ratio} is set to 0.7.
We set the attack budget to 1\% tokens for Gigaword and 5\% tokens for CNN/DM, based on the consideration of attacking performance and semantic consistency.
We use the original dataset plus the augmented cases generated by SummAttacker as our training dataset, where we also randomly add 30\% natural human errors to improve the understanding of noises.
The training process takes about 8 hours and 4 hours for CNN/DM and Gigaword.
% More details can be found in Appendix~\ref{implemen_appendix}.

\subsection{Evaluation Metrics}
We first evaluate models using standard ROUGE F1~\cite{lin2004rouge}. 
ROUGE-1, ROUGE-2, and ROUGE-L refer to the matches of unigrams, bigrams, and the longest common subsequence, respectively.
We use BERTScore \cite{zhang2019bertscore} to calculate similarities between the summaries. 
We further evaluate our approach with the factual consistency metric, QuestEval \cite{scialom2021questeval} following \citet{chentowards}.
It measures to which extent a summary provides sufficient information to answer questions posed on its document.
\textcolor{black}{QuestEval considers not only factual information in the generated summary, but also the information from its source text, and then gives a weighted F1 score.}

\section{Experimental Results}

\begin{table}[tb]\setlength{\tabcolsep}{2pt}\small
    \centering
    \begin{tabular}{ccccc|cc}
        \toprule
        \multicolumn{3}{c}{\textbf{Dataset}}   & \textbf{Semantic } & \textbf{Grammar }& \textbf{Similarity}\\
        \midrule
        \multirow{2}*{\textbf{Gigaword}} & &Original  & 4.4  & 4.7 &-  \\
        & & Adversarial &4.1 & 4.5 & 0.96\\
        \midrule
        \multirow{2}*{\textbf{CNN/DM}} && Original & 4.4 & 4.6&-  \\
        & & Adversarial &4.0& 4.2 & 0.94 \\
        \bottomrule

    \end{tabular}
    \caption{Human  and automatic evaluation of the adversarial samples from SummAttacker, as well as  the original samples for taking a reference. 
    }

    \label{tab:humaneval}
\end{table}

\begin{table*}[htb]
\small
\centering
\resizebox{0.8\textwidth}{!}{
\begin{tabular}{c|c|ccc|cccc}
\toprule[1pt]
\multirow{2}{*}{Dataset} & \multirow{2}{*}{Model} & \multicolumn{3}{c|}{Traditional Metric}          & \multicolumn{4}{c}{Advanced Metric} \\
                         &                        & ROUGE-1 & ROUGE-2 & \multicolumn{1}{c|}{ROUGE-L} & BERTScore & QE(R)   & QE(P)  &QE(F1)   \\ 
                      \hline   \multirow{5}{*}{Gigaword}   
                         
                        & BART & 35.23 & 15.64 & 32.52 & 87.33 & 22.42 & 22.32 & 22.37 \\
 & ProphetNet & 35.56 & 15.87 & 32.79 & 88.45 & 23.48 & 23.76 & 23.62\\
 & R3F & 35.69 & 16.29 & 32.91 & 88.60 & 23.05 & 23.79 & 23.42 \\
 & SSTIA & 36.55 & 16.90 & 33.25 & 88.72 & 23.52 & 24.01 & 23.76 \\ 
 \cmidrule{2-9}
 &DASum & \textbf{38.15} & \textbf{18.53}	&\textbf{35.31} & \textbf{88.90}	& \textbf{27.39}	&\textbf{28.95} &\textbf{28.17}
                         \\
                          &DASum w/o  $\mathcal{L}_e$ & 36.71               & 18.17	&34.01 &88.61 & 24.89 &26.63& 25.76 \\
                           &DASum w/o $ \mathcal{L}_d$ & 37.36	 &18.31	&  34.64 & 88.71            & 24.64	& 26.93	&25.79      
                         \\
                          &DASum w/o $\mathcal{L}_c$  & 37.21	 &18.30	&  34.32 &    88.64     & 25.56	& 26.19	&25.87       
                         
                         \\
                          
                         \hline
\multirow{7}{*}{CNN/DM}   
              & BART & 36.45 & 12.29 & 33.36 & 87.23 & 22.05 & 17.47 & 19.76\\
 & ProphetNet & 36.98 & 12.68 & 33.8 & 87.33 & 22.28 & 17.43 & 19.85 \\
 & R3F &37.28 & 12.98 & 34.83 & 87.59 & 22.14 & 17.88 & 20.01\\
 & SSTIA & 37.49 & 13.05 & 35.15 & 87.69 & 22.46 & 17.96 & 20.21\\
  \cmidrule{2-9} 
 &DASum & \textbf{42.17} & \textbf{18.06} & \textbf{39.08} & \textbf{88.90} & \textbf{28.66} & \textbf{25.62} & \textbf{27.14}\\

 \bottomrule[1pt]
\end{tabular}
}
\caption{Performance of baselines and our model DASum on perturbed inputs by SummAttacker (the attack budget is 1\% and 5\% tokens in Gigaword and CNN/DM datasets respectively. 
Numbers in \textbf{bold} mean that the improvement to the best baseline is statistically significant (a two-tailed paired t-test with p-value \textless 0.05).
 }
\label{table:combine}
\end{table*}

\subsection{SummAttacker Evaluation}

Before \textcolor{black}{reporting the summarization performance boosted by our proposed dual augmentation strategy, }
we first set up human and automatic metrics to evaluate the quality of the generated adversarial augmentation cases.
For human evaluation, we ask annotators to score the semantic and grammar correctness of the generated adversarial and original sequences, scoring from 1-5 following \citet{jin2019bert} and \citet{li2020bert}.
% Then we ask annotators to classify each example in the shuffled mix of the original and adversarial sentences. 
We randomly select 100 samples of both original and adversarial samples for human judges. 
Each task is completed by three Ph.D. students.
% For label prediction, we use the majority class as the predicted label for label prediction and use the average score for semantic and grammatical checks.
For automatic metric, following \citet{li2020bert}, we use Universal Sentence Encoder \cite{cer2018universal} to measure the semantic similarity between the adversarial and the original documents. 
% \textcolor{black}{Table~\ref{tab:humaneval} presents the averaged metric scores over 100 samples.} 
% We also examine the influence of the candidate number $K$.

As shown in Table~\ref{tab:humaneval}, the adversarial samples' semantic and grammatical scores are reasonably close to those of the original samples.
The scores are generally higher on Gigaword dataset than CNN/DM.
This corresponds to the setting that the number of attacked words is larger on CNN/DM dataset.
The kappa statistics are 0.54 and 0.48 for semantic and grammar respectively, indicating moderate agreements between annotators.
For the automatic evaluation, the high semantic similarity demonstrates the consistency between the original and attacked documents.

\begin{figure}[tb]
\centering
\includegraphics[scale=0.25]{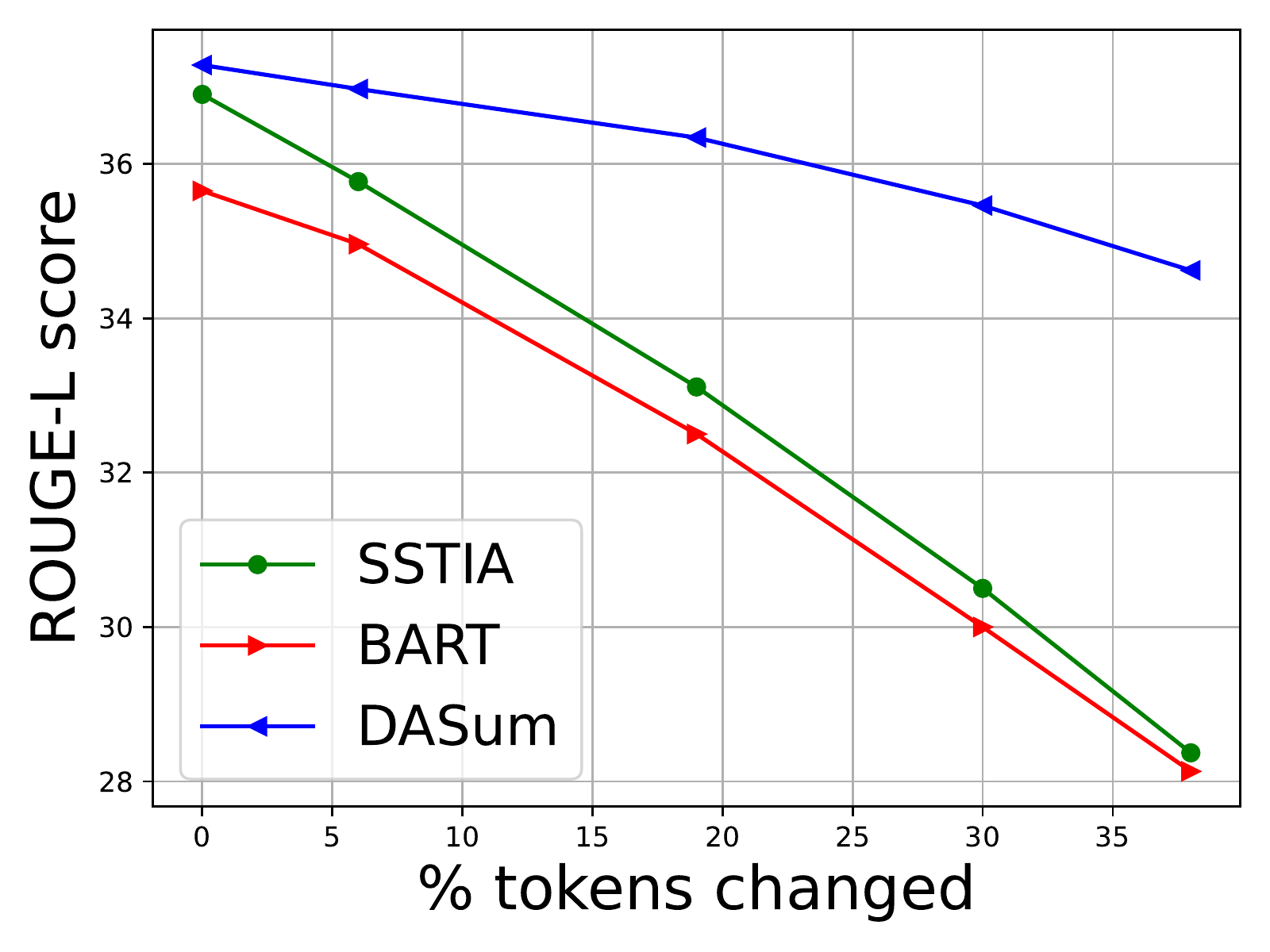}
\caption{
    The \textcolor{black}{  impact of noise on the performance of summarization models on Gigaword. While SSTIA and BART show significant drops in all metrics, our DASum has a robust performance.   } 
	The noise here consists of multiple human errors (typos, misspellings, etc.)
}

\label{fig:contras}
\end{figure}

\definecolor{Gray}{gray}{0.96}
\begin{table*}[tb]
\resizebox{\textwidth}{!}{
\centering
% \footnotesize
\small
\begin{tabular}{p{7.2cm}p{2.6cm}p{2.6cm}p{2.6cm}p{2.6cm}}
\toprule
 \multicolumn{1}{c}{Attacked Document \& Reference 
}&\multicolumn{1}{c}{SSTIA on clean input}&\multicolumn{1}{c}{SSTIA on attacked input}&\multicolumn{1}{c}{DASum on clean input }&\multicolumn{1}{c}{DASum on attacked input}\\ \midrule
 \rowcolor{Gray}
\textbf{Doc:} overcrowding and lick of illumination at exit popints at konkola stadiom in UNK province of zambia were among tu major lapses that lead to a stampede resulting in the dieth of \#\#  sokker fun afrer ana africa coop... 
% between zambia und Congo UNK onf June \#, thi year, sports minister gabriel UNK told the parlament Thurday.
& \hlc[crimsonglory!20]{overcrowding blamed} for stampede in zambia & \#\# zambian soccer fans injured in stampede. &\hlc[yellow!40]{overcrowding blamed} for soccer stampede in zambia
  &\hlc[yellow!40]{overcrowding blamed} for soccer stampede in zambia
 \\
  \rowcolor{Gray}
 \textbf{Ref:} overcrowding lack of illumination leads to stampede in zambia: investigation &&&&\\

\textbf{Doc:} philippine president fidel ramos, who was hospitalized for the second time in \#\# days over the weekend, may need heart surgery, his \hlc[bleudefrance!40]{spokesman ($\rightarrow$spokesperson)} said.
\textbf{Ref:} philippine president hospitalized may need heart surgery
 &\hlc[crimsonglory!20]{philippine president} may need heart surgery&
ramos may need heart surgery 
  & \hlc[yellow!40]{philippine president} may need heart surgery spokesman say&
\hlc[yellow!40]{philippine president} ramos may need heart surgery\\

  \rowcolor{Gray}

\textbf{Doc:} gusty winds pushed a \hlc[bleudefrance!40]{wildfire ($\rightarrow$bonfire)} closer to sun valley resort 's ski area, while hundreds more homes were ordered evacuated in the valley below."
\textbf{Ref:} gusty winds whip idaho wildfire near sun valley ski area ; hundreds more homes evacuated
 &hundreds more homes evacuated as wildfire threatens ski resort &
hundreds more homes evacuated as winds push bonfire closer to \hlc[crimsonglory!20]{california ski resort}
  & winds push wildfire closer \hlc[yellow!40]{to sun valley ski area} &
winds push wildfire closer \hlc[yellow!40]{to sun valley ski area}\\

  \bottomrule
\end{tabular}}
\caption{Comparisons of summaries generated by baseline models and our method on the noisy document (the first row) and attacked document (last two rows).
The \hlc[crimsonglory!20]{missing information} or \hlc[crimsonglory!20]{inconsistent information} caused by perturbations on the baseline model and the \hlc[yellow!40]{consistent information} given by our model is highlighted.}
\label{tab:ptlm-bias-mini}
\end{table*}

We also study the influence of the candidate number $K$  in SummAttacker.
In Fig.~\ref{fig:candidate}, \textcolor{black}{all models  perform worse when the input document is perturbed by SummAttacker with a larger $K$, }
%the attacking performance gets better as the candidate number $K$ increases, 
since  a better  replacement word $w'_i$ can be found in a larger search space.
From the viewpoint of generating adversarial samples, it is not worth using a  large $K$, because  the time and memory complexity increase with $K$ as well. 
%We also find that when $K$ is large, continuing to increase $K$ is not worth while since the time and memory complexity increase. 
Thus, we use $K$=10 in our setting.
%Intuitively, a larger $K$ would result in less semantic similarity. 
%The candidates are all logical and semantically coherent with the original sentence, according to the Universal Sentence Encoder semantic metric, where the score is kept in a steady range (similarities drop less than 1\%).
% Performance in other metrics is shown in Appendix.
% we can see that the summarization performance gets worse as the candidate number $K$ increases, and the influence gets smaller when $K$ is large.
% Hence, we set the default number as 10.

\subsection{Robustness Evaluation}
\label{robustevaluate}
\textcolor{black}{We next report the evaluation results of summarization models when the input documents are perturbed by natural human errors (noise) and synonym substitutions (based on SummAttacker). }
\textbf{Robustness on Noisy Datasets.}
Humans make mistakes when typing or spelling words, but they  have the capability of comprehensive reading to understand the document without being interrupted by  such noises.
Thus, we first examine the robustness of the recent summarization models against natural human errors.
Since we do not have access to a summarization test set with natural noise,
we use the look-up table of possible lexical replacements \cite{belinkov2017synthetic}, which collects naturally occurring errors (typos, misspellings, etc.).
We replace different percentages of words in the Gigaword test set with an error if one exists in the look-up table.
% When there is more than one possible replacement to choose we sample uniformly. 
We show the  performance of classic baseline BART, augmentation-based model SSTIA, and our model in Fig.~\ref{fig:contras}.
Both baseline models suffer a significant drop in all metrics when evaluated on texts consisting of different percentages of noise. 
Our DASum model is more robust and drops the least in all four metrics compared with baselines.
We also give an example in the first row in Table~\ref{tab:ptlm-bias-mini}. 
Humans are quite good at understanding such scrambled texts, 
 \textcolor{black}{whereas existing summarization models are still vulnerable to slight perturbations and then fail to capture the gist of the input document, due to the lack of robustness enhancement training}.
%Considering that the summarization task is to capture the gist of the document instead of comprehending a specific number of words, the degradation in summarization quality is especially severe.

\begin{figure}[tb]%
    \centering
    \subfigure{{\includegraphics[width=3.6cm]{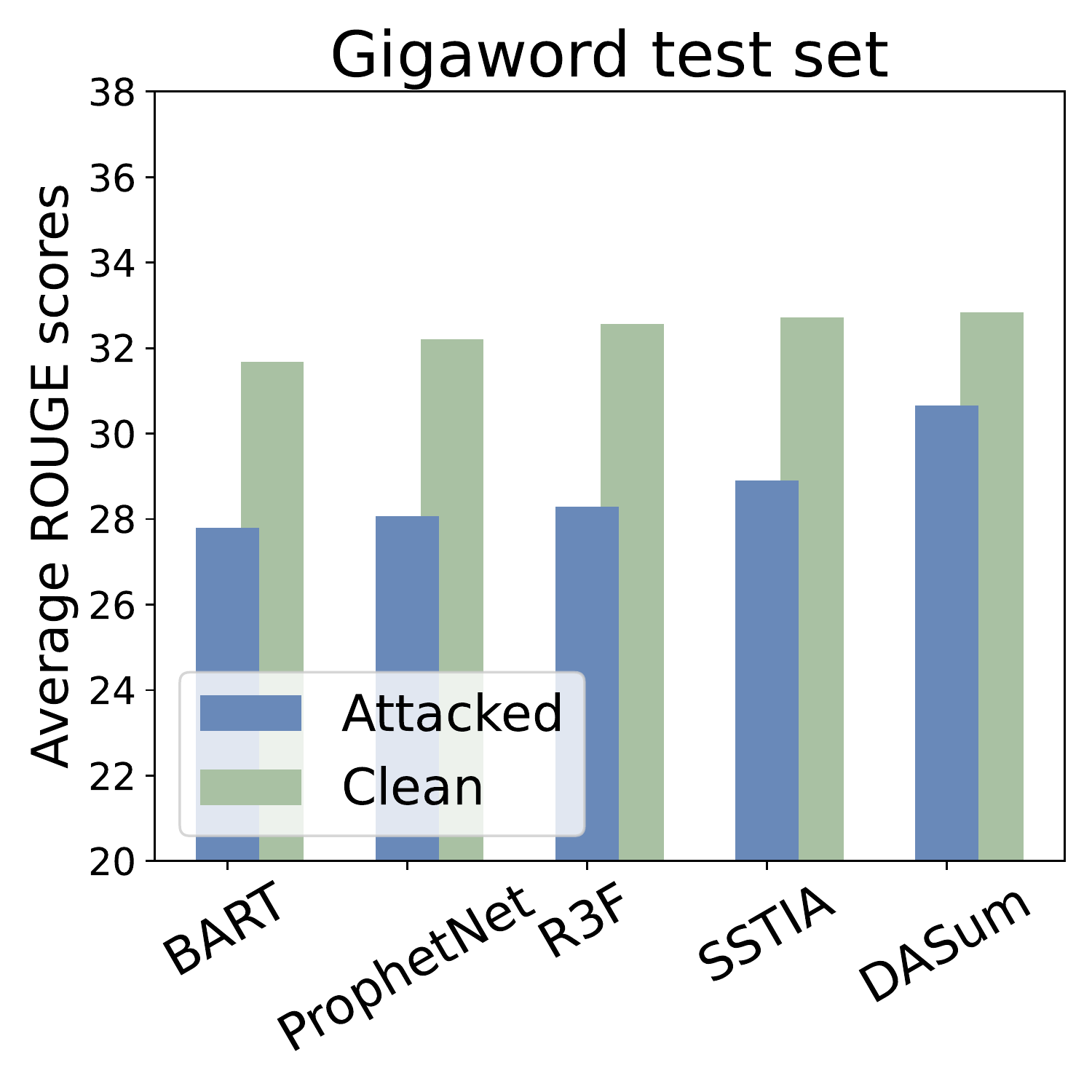} }}%
    \subfigure{{\includegraphics[width=3.6cm]{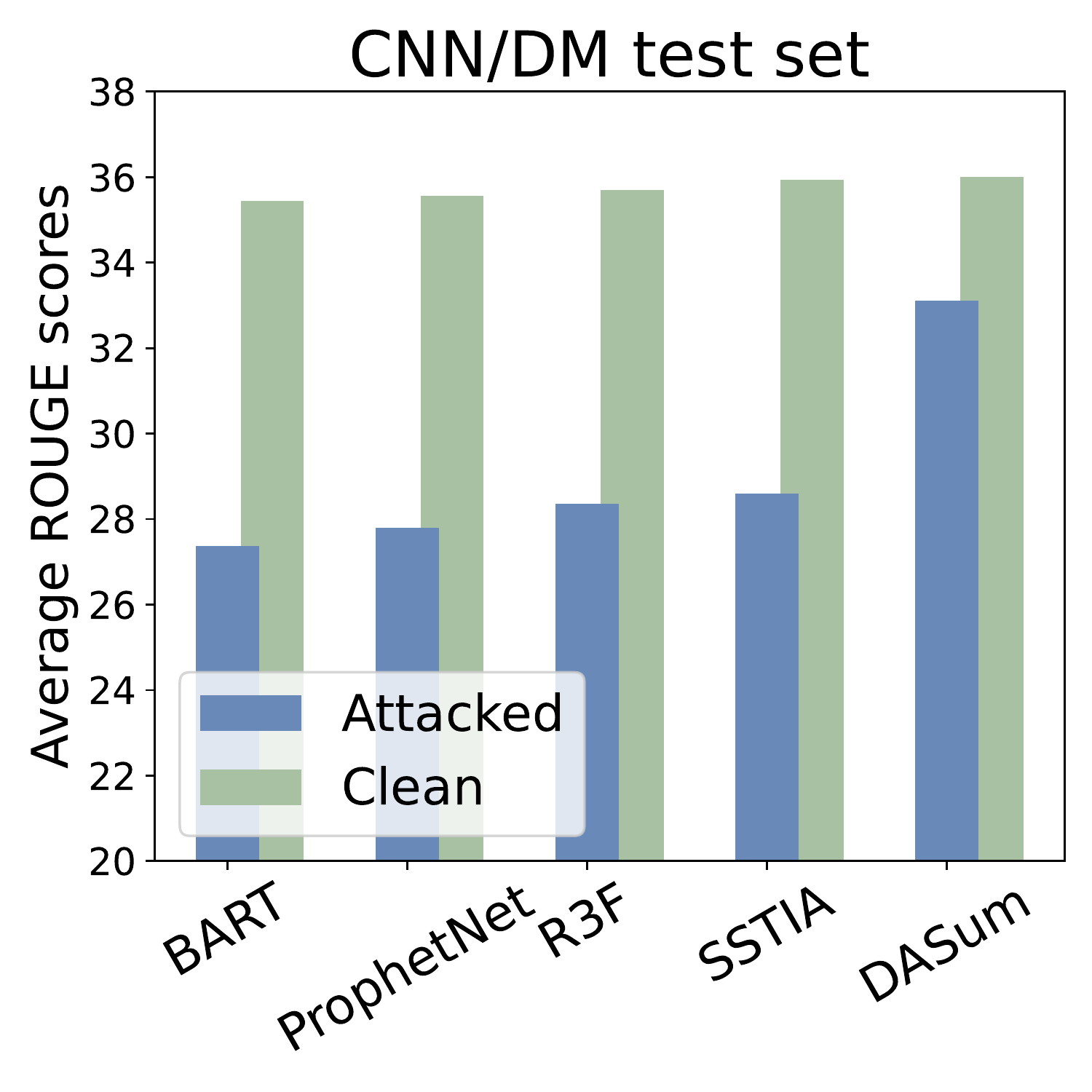} }}%
    \caption{Performance   of baselines and our model on attacked and clean Gigaword and CNN/DM test set.}%
    \label{f2}%
\end{figure}

\textbf{Robustness on  Datasets Perturbed by Adversarial Attacks.}
We next examine the robustness of  summarization models on the test datasets  perturbed by adversarial attacks.
% Concretely, the test set is the combination of the original clean test set and the attacked set, since in real-world applications, the inputs are always diverse and mixed.
For the Gigaword dataset, we set attack budget $\epsilon$ to be  only 1 word (1\% tokens), and for CNN/DM we set $\epsilon$ to be 5\% tokens of the input document.

\textcolor{black}{The comparison of performance on attacked and clean datasets is shown in Fig.\ref{f2}.}
It can be seen that despite the perturbation being only on a few words, all four baselines suffer a significant drop in performance compared with their performance on the clean test set. 
Specifically, the ROUGE-1 score of the latest SSTIA model drops by 4.01 on Gigaword, and the average ROUGE score drops by 7.33 for R3F model on CNN/DM dataset.
This highlights the vulnerability of the existing summarization models and also demonstrates the effectiveness of our attacking model.
Nevertheless, the drop percentage of our model is the least compared with other baselines in all metrics. 
Specifically, our model drops the least with only 2.22 and 0.28 decreases in ROUGE-2 and BERTScore metrics, respectively, on the Gigaword dataset.
We show the detailed performance on attacked set in Table~\ref{table:combine}.
Our model outperforms baselines on two datasets in most metrics.
Besides, we also observe that \textcolor{black}{the summarization models of short documents are more vulnerable than those of long documents.}
One potential reason is that the summarization model is more dependent on each input word when the input is shorter.
When the input is longer, the importance of each word decreases, since the model can resort to other sources to generate summaries.

\textcolor{black}{\textbf{Ablation Study.}}
We first investigate the influence of \textbf{\textit{input space augmentation}}.
As shown in Table~\ref{table:combine}, without the $\mathcal{L}_e$ loss, the performance drops the most.
We also conduct diversity analysis on the inputs after augmentation, corresponding to the vulnerability discussion in $\S$\ref{dual}.
The ratio of uncommon words compared with the original common words increases by 30\%, which directly verifies our assumption that introducing variations in the training dataset improves the robustness of the summarization model.
Next, we study the effect of \textit{\textbf{latent space augmentation}}.
Specifically, the ROUGE-1 score of extractive summarization drops by 0.79 after the $\mathcal{L}_d$  is removed. 
This indicates that the model benefits from hidden states with more diversity in the training process. 
\textcolor{black}{In addition, we compare the decoder hidden states of DASum with that of BART in Fig.\ref{f1}.  The deviation of paired original and attacked hidden states in DASum is effectively reduced ($\overline{E}$ drops from 2.5 to 1.9 in the last layer).}
Thirdly, the performance of DASum w/o $\mathcal{L}_c$ shows that \textit{dual consistency} can also help improve robustness.
We also note that DASum is always more robust than the other two baselines, in regard to different attacking settings in Fig.\ref{fig:contras}.

\section{Conclusion}

In this paper, we investigate the robustness problem in   the summarization task, which has not been well-studied before.
We first come up with a SummAttacker, which slightly perturb    the input documents in benchmark test datasets, and causes a significant performance drop for the recent summarization models.
Correspondingly, we propose a dual data augmentation method for improving the robustness,
which generates discrete and virtual training cases in the same meaning but with various expression formats. 
Experimental results show that our model outperforms strong baselines.
% In the future, we would like to address the robustness of summarization models at the sentence level and document level.

\section*{Limitations}
We discuss the limitations of our framework as follows:

(1) In this paper, we take an initial step on the robustness of the summarization system by focusing on word-level perturbations in the input document.
However, in practice, the robustness of the summarization models is reflected in many other aspects.
For example, the summarization performance towards sentence-level or document-level perturbations is also a kind of robustness.
% We give a rough evaluation on sentence-level robustness in Appendix~\ref{sent_appendix}, and we leave the detailed evaluation of such robustness and improvement methods for future work.

(2) Although DASum greatly improves the generation quality compared with other augmentation-based models, it requires more computational resources with respect to the augmented dataset construction process. 
For large-scale datasets with long text (e.g., BigPatent~\cite{sharma2019bigpatent}), it is worth considering the time complexity of Transformer architecture.

\section*{Acknowledgments}

We would like to thank the anonymous reviewers for their constructive comments. 
The work was supported by King Abdullah University of Science and Technology (KAUST) through grant awards FCC/1/1976-44-01, FCC/1/1976-45-01, REI/1/5234-01-01,  RGC/3/4816-01-01, and RGC/3/4816-01-01;.

\bibliography{custom}

\end{document}